\relax
\documentclass[letterpaper]{article} %
\usepackage{aaai21}  %
\usepackage{times}  %
\usepackage{helvet} %
\usepackage{courier}  %
\usepackage[hyphens]{url}  %
\usepackage{graphicx} %
\usepackage{latexsym,comment}
\usepackage{amsfonts}
\usepackage{natbib}  %
\usepackage{caption} %

\usepackage{amsmath}

\newcommand{\knn}{\(k\)NN}
\newcommand{\Knn}{\(K\)NN}
\DeclareMathOperator*{\argmax}{arg\,max}
\usepackage{microtype,booktabs,multirow}
\usepackage{array}
\newcolumntype{P}[1]{>{\centering\arraybackslash}p{#1}}
\usepackage{amsmath}
\usepackage{xcolor}
\usepackage{booktabs}
\usepackage{enumerate}
\usepackage{multirow,siunitx}
\usepackage{arydshln}
\usepackage[english]{babel}
\usepackage[autostyle, english=american]{csquotes}
\MakeOuterQuote{"}
\usepackage{soul}
\usepackage{mwe}
\usepackage{subfigure}

\title{Explaining and Improving Model Behavior with \(k\) Nearest Neighbor Representations}
\author{Nazneen Fatema Rajani\quad\quad Ben Krause\quad \quad Wenpeng Yin \quad\quad Tong Niu \\ Richard Socher \quad\quad Caiming Xiong \\}

\affiliations{Salesforce Research \\ \texttt{nazneen.rajani@salesforce.com}}

\begin{document}
\maketitle
\begin{abstract}
 Interpretability techniques in NLP have mainly focused on understanding individual predictions using attention visualization or gradient-based saliency maps over tokens. We propose using \(k\) nearest neighbor (\knn)~representations to identify training examples responsible for a model's predictions and obtain a corpus-level understanding of the model's behavior. Apart from interpretability, we show that \knn~representations are effective at uncovering learned spurious associations, identifying mislabeled examples, and improving the fine-tuned model's performance. We focus on Natural Language Inference (NLI) as a case study and experiment with multiple datasets. 
 Our method deploys backoff to \knn~for BERT and RoBERTa on examples with low model confidence without any update to the model parameters. Our results indicate that the \knn~approach makes the finetuned model more robust to adversarial inputs.
\end{abstract}
\section{Introduction}
\label{sec:introduction}

Deep learning models are notoriously opaque, leading to a tremendous amount of research on the inscrutability of these so-called black-boxes. Prior interpretability techniques for NLP models have focused on explaining individual predictions by using gradient-based saliency maps over the input text~\citep{lei2016rationalizing,ribeiro2018anchors,bastings-etal-2019-interpretable} or interpreting attention~\citep{brunner2019validity, pruthi2019learning}. These methods are limited to understanding model behavior for example-specific predictions. 

In this work, we deploy \(k\) Nearest Neighbors (\knn) over a model's hidden representations to identify training examples closest to a given evaluation example. By examining the retrieved representations in the context of the evaluation example, we obtain a dataset-level understanding of the model behavior. Taking the NLI  problem as an example (Figure~\ref{fig:knn}), we identify the confidence interval where \knn~performs better than the model itself on a held-out validation set. During inference, based on the model's confidence score we either go forward with the model's prediction or backoff to its \knn~prediction. Our implementation of \knn~can be used with any underlying neural classification models and it brings three main advantages:

\begin{figure}[t!]
\centering
\includegraphics[width=0.48\textwidth]{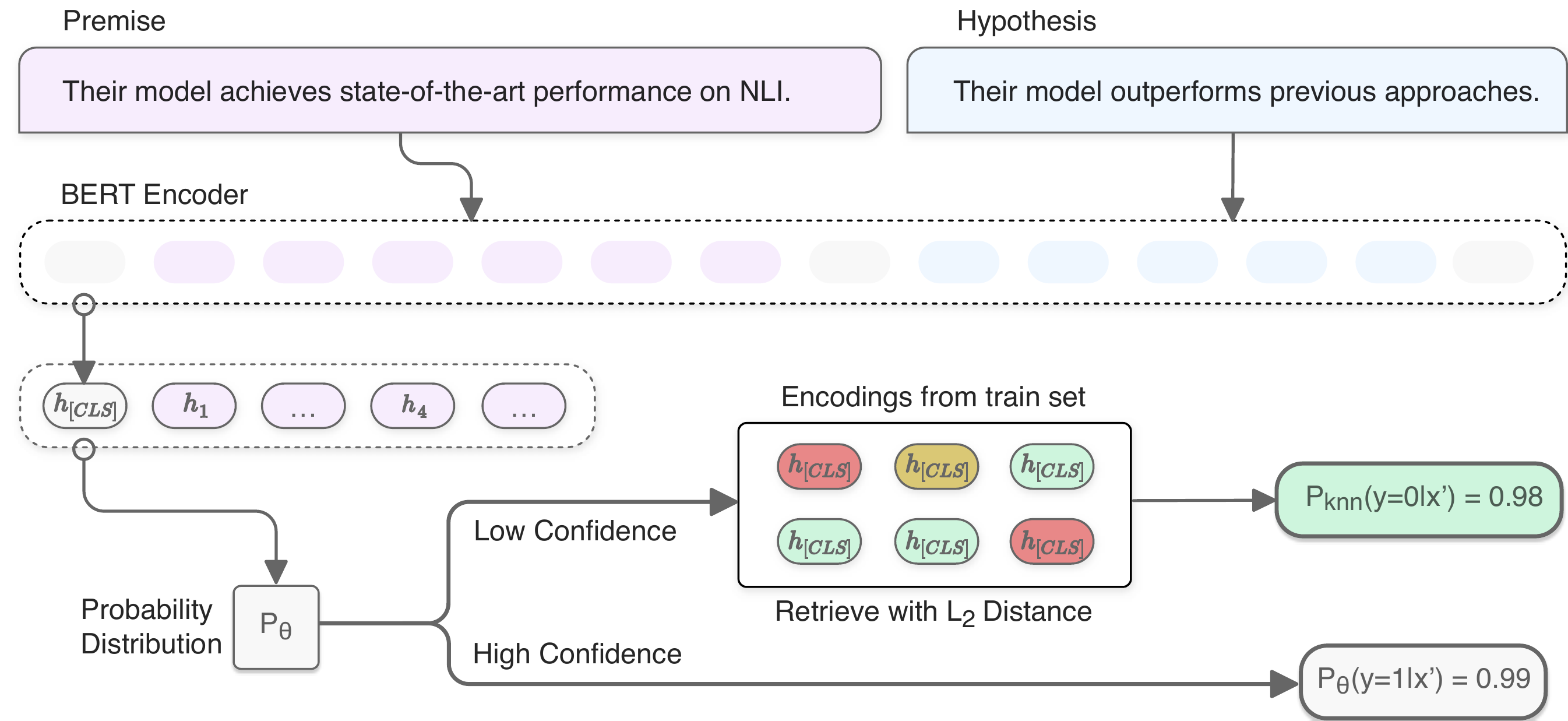}
\caption{\small Overview of our proposed approach on using \knn~as backoff to improve fine-tuned classification models.}
\label{fig:knn}
\end{figure}

\textbullet\enspace \textbf{Explaining model behavior}. Our \knn~approach is able to explain a model's prediction by tracing it back to the training examples responsible for that prediction. Although it sounds similar with influence functions \citep{koh2017understanding}, our approach is much simpler because it does not access the model parameters and is approximately $300\times$ faster on small datasets ($<2K$ examples) and the gap increases on bigger datasets.

\textbullet\enspace \textbf{Uncovering  spurious associations}. 
We observe that our \knn~model can learn more fine-grained decision boundaries due to its added non-linearity, which can make it more robust to certain kinds of spurious correlations in the training data. We leverage this robustness for studying where models go wrong, and demonstrate how retrieving the nearest neighbors of a model's misclassified examples can reveal artifacts and spurious correlations. From our analysis, we also observed that \knn~of misclassified test examples can often retrieve mislabeled examples, which makes this approach applicable to fixing mislabeled ground truth examples in training sets.

\textbullet\enspace \textbf{Improving model predictions}. 
Along with providing a lens into understanding model behavior, we propose an approach that interpolates model predictions with \knn~for classification by identifying a more robust boundary between classes. We are the first to demonstrate this using \knn~for both interpretability and for improving model performance.

In summary, we propose a \knn~framework that uses hidden representations of fine-tuned models to explain their underlying behaviors, unveil learned spurious correlations and further improve the model predictions.  Through \knn~we shed light on some of the existing problems in NLI datasets and deep learning models, and suggest some evaluation prescriptions based on our findings. To our knowledge, this is the first work that uses \knn~in the representation space of deep neural models for both understanding and improving model behavior.

\section{Related work}
\label{sect:related work}
Our work is related to three main areas of research: approaches for interpretability in NLP, retrieval-based methods, and work on dataset artifacts and spurious associations.

\paragraph{Interpretability methods in NLP.}
Our approach of using \knn~to retrieve nearest training examples as an interpretability technique is model-agnostic. LIME \cite{ribeiro2016should} is another model-agnostic method that interprets models by perturbing inputs and fitting a classifier in the locality of the model's predictions for those perturbations. It has been extended to identifying artifacts and biases in information retrieval~\citep{singh2018exs}.~\citet{alvarez2017causal} probe the causal structure in sequence generation tasks by perturbing the input and studying its effect on the output. The explanation consists of tokens that are causally related in each input-output pair. 

Our method is closely related to the work on influence functions~\citep{koh2017understanding} which has been recently extended by~\citet{han2020explaining} to neural text classifiers. Influence function are accurate for convex  only when the model is strictly convex and so the authors approximation to the influence functions on BERT \cite{devlin2018bert}. The authors evaluate the effectiveness by comparing them with gradient-based saliency methods for interpretability. Our approach uses FAISS~\cite{johnson2019billion} to create a cache once and so is computationally efficient and scalable to very large datasets that deep learning models rely on.

Interpreting attention as a form of explanation for neural NLP models has been a topic of debate~\citep{jain2019attention,serrano2019attention,wiegreffe2019attention}.~\citet{zhong2019fine} train attention to directly attend to human rationales and obtain more faithful explanations. Recent work has manipulated attention in Transformer \cite{DBLPVaswaniSPUJGKP17} models in an attempt to make it more interpretable~\citep{pruthi2019learning,brunner2019validity}. 
Gradient-based saliency maps are faithful by construction and hence have been applied in various forms to neural models for interpretability~\citep{lei2016rationalizing,ribeiro2018anchors,chang2019game,bastings-etal-2019-interpretable}.

\paragraph{Retrieval-based approaches.}
Many language generation methods use retrieval-based techniques at test time.~\citet{weston2018retrieve} improve a dialogue response generation model by retrieving and refining similar instances from the training set.~\citet{gu2018search} propose a Neural Machine Translation approach that fuses information from the given source sentence and a set of retrieved training sentence pairs. 
The neural cache \citep{grave2017improving} and related unbounded neural cache \citep{grave2017unbounded} retrieve similar instances from the sequence history to better adapt language models to the context. The recently introduced \knn~language model (\knn-LM) extends on existing pre-trained language models by interpolating the next word distribution with a \(k\)NN model~\citep{khandelwal2019generalization}. The combination is a meta-learner that can be effectively tuned for memorizing and retrieving rare long-tail patterns. Our work adapts the \(k\)NN-LM to text classification by using the input example as context, in a similar way to how deep-\knn~\citep{papernot2018deep} applies \knn~over neural network representations to the classification of images. We show that our approach outperforms fine-tuned models like BERT and RoBERTa \cite{liu2019roberta} and is particularly good as a backoff for a model's less confident examples.

In addition to \textit{k}NN-LM style approaches that interpolate the next word distribution, there has also been work leveraging the \textit{k}NN training examples of a test instance to fine-tune a pretrained model~\cite{zhao2020reinforced}, which avoids the need to create new model architecture and train from scratch. For example,~\cite{li-etal-2018-one} search for similar sentence pairs from the training corpus for each testing sentence, which are then used to finetune the model.

\paragraph{Dataset artifacts and spurious associations.}
Neural models have been very successful on various NLP tasks but a deeper understanding of these models has revealed that these models tend to exploit dataset artifacts. In Natural Language Inference (NLI), the hypothesis-only baseline is known to significantly outperform the majority baseline on both the SNLI~\citep{snli:emnlp2015} and MNLI~\citep{N18-1101} datasets~\citep{gururangan2018annotation,poliak2018hypothesis}. Pre-trained models like BERT rely on spurious syntactic heuristics such as lexical overlap for NLI~\citep{mccoy2019right}. Various forms of data augmentations and evaluations for robustness have been proposed to overcome these limitations. ~\citet{kaushik2019learning} use counterfactual augmentation for SNLI to alleviate the bias from annotation artifacts. Their findings show that training on counterfactuals makes models more robust. We use \knn~to study the shift in representations before and after the augmentation in terms of spurious signals. Our findings suggest that although counterfactual augmentation does help with reducing the effects of prominent artifacts like lexical overlap, it also inadvertently introduces new artifacts not present in the original dataset.

\section{Method}
\label{sec:model}
Figure~\ref{fig:knn} shows an overview of our proposed approach that leverages \knn~to improve an underlying classification model. Our implementation of \knn~relies on caching the model's output hidden representation of every training sequence. During inference, we query the cache to retrieve the nearest neighbors for a given example and make a prediction based on the weighted distance for each class.

\subsection{Learning hidden representations}
Our approach to \knn~over hidden representations closely relates to \knn-LM implemented by \citet{khandelwal2019generalization} but has an extra normalization step over hidden states and a different method for interpolating \knn~and neural networks for classification. 

Our implementation assumes that we have a training set of sequences where each sequence $x^i$ is paired with a target label $y^i$. Our algorithm maps each $x^i$ to a hidden representation vector $h^i  \in \mathbb{R}^d$, where $d$ is the hidden state dimension, using function $f_{\theta}$ defined by a neural network with parameters $\theta$: 
\begin{equation*}
h^i = f_{\theta}(x^i)
\end{equation*}
In this work, these hidden states are obtained from BERT or RoBERTa systems.
They are crucial in determining the performance of our method. We found that the \texttt{[CLS]} representation of the last layer of BERT or RoBERTa performs the best. As prescribed in \cite{reimers2019sentence}, we also experimented with other pooling strategies like mean and max pooling of hidden representations across all time steps but the improvements were smaller compared to the \texttt{[CLS]} representation. We also evaluated using the best layers identified for capturing input semantics in generation for various transformer models~\citep{zhang2019bertscore}. However, we found that the last layer worked best and so all our models use the \texttt{[CLS]} representation of the last layer.

Those hidden representations can be collected and cached with one forward pass through the training set.
For scaling to larger data sets, FAISS implements data structures for storing the cache that allows for faster \knn~lookup and reduces memory usage. We then apply dataset-wise batch normalization \citep{ioffe2015batch} to the hidden state vectors with means $\mu \in \mathbb{R}^d$ and standard deviations $\sigma \in \mathbb{R}^d$ over the hidden states, and obtain normalized hidden states using

\begin{equation*}
\tilde{h}^i =  \frac{h^i - \mu}{\sigma + \epsilon},
\end{equation*}
with a small $\epsilon$ for numerical stability.
\noindent
We map each test sequence $x'$ to a fixed-size hidden state $\tilde{h}'$ with
\begin{equation*}
\tilde{h}' =  \frac{f_{\theta}(x')- \mu}{\sigma + \epsilon}.
\end{equation*}

 When the training set is large, it is possible to estimate means and variances from a subset of training sequences.

\vskip 1em
\subsection{\(K\) nearest neighbors over hidden representations}
For $k$ nearest neighbors, we first locate the set of indices $K$ for $i$ which leads to the smallest $L_2$ distances given by $d^i = ||\tilde{h}'-\tilde{h}^i||^2$. Next, we compute the weighted \knn~probability scores $w_{knn}(x^i,x')$ with a softmax over negative distances given by

\[
    w_{knn}(x^i,x')= 
\begin{cases}
    \frac{\exp(-d^i/T)}{\sum_{j \in K} \exp(-d^j/T)},& \text{if } i \in K\\
    0,              & \text{otherwise.}
\end{cases}
\]
where $T$ is a temperature hyper-parameter that controls the sharpness of the softmax. The probability distribution over labels for the test sequence, $p_{knn}(y|x')$ is then given by 
\begin{equation*}
p_{knn}(y|x') = \sum_{j \in K} w_{knn}(x^j,x') \times e_{(y^j)}
\end{equation*}
where $e_{(y^j)}$ is a one-hot encoding of $y^j$ equal to one at the index of $y^j$ and zero at the index of all other labels. $p_{knn}(y|x')$ can be used directly as a classifier or interpolated with the base neural network probability distribution $p_{\theta}(y|x')$ in various ways.

In this work, we backoff to a \knn~classifier depending on pre-defined criteria, such as when the model is less-confident in its predictions or if the model is known to perform poorly on certain types of inputs. Given some threshold hyper-parameter $\tau$, our classifier prediction is given by  
\[
y=
\begin{cases}
    \argmax p_{\theta}(y|x'),& \text{if }  \max p_{\theta}(y|x') > \tau \\
     \argmax  p_{knn}(y|x'),              & \text{otherwise}
\end{cases}
\]
The hyper-parameters $\tau$ and $T$ are determined based on each model and the validation set. We tune the value of $k$ on the validation set of each dataset, and use the same value for all models trained on that dataset.

\section{Experiments}
\label{sec:experimental setup}
\vskip 1em
\subsection{Datasets}
\label{subsec:datasets}
We demonstrate the effectiveness of \knn~on the natural language inference (NLI) task as a case study. The input to the NLI model is a pair of sentences -- the premise and the hypothesis, and the task is to predict the relationship between the two sentences. 
The possible labels are `entailment',`contradiction', or `neutral'. This problem requires complex semantic reasoning of underlying models~\citep{dagan2005pascal}.

Apart from the vanilla version of the task, we also compare to augmented and adversarial versions of the original datasets to gain a deeper understanding of how the model behavior changes.

\textbullet\enspace \textbf{SNLI:}
The Stanford Natural Language Inference (SNLI)~\citep{snli:emnlp2015} dataset is a widely used corpus for the NLI task. Recently, ~\citet{kaushik2019learning} released a counterfactually augmented version of the SNLI. The new corpus consists of a very small sample of the original dataset ($0.3\%$) called the original split. The original split is augmented with counterfactuals by asking crowdworkers to make minimum changes in the original example that would flip the label. 
This leads to three more splits -- the revised premise wherein only the premise is augmented, the revised hypothesis wherein only the hypothesis is augmented or the combined that consists of both premise and hypothesis augmentations along with the original sentence pairs. 

 We use the original and combined splits (refered to as augmented split) in our experiments that have training data sizes of 1666 and 8330 respectively. For validation and testing on the original split, we use the SNLI validation and test sets from~\citet{snli:emnlp2015} with sizes 9842 and 9824, respectively. For the combined split, we validate and test on the combined validation and test sets with sizes 1000 and 2000, respectively.

\vskip 1em
\textbullet\enspace \textbf{ANLI:}
To overcome the problem of models exploiting spurious statistical patterns in NLI datasets,~\citet{DBLPabs191014599}
released the Adversarial NLI (ANLI) dataset.~\footnote{Available at ~\url{https://www.adversarialnli.com/}} ANLI  is a large-scale NLI dataset collected via an iterative, adversarial human-and-model-in-the-loop procedure. In each round, a best-performing model from the previous round is present, then human annotators are asked to write ``hard'' examples the model misclassified. They always choose multi-sentence paragraphs as premises and write single sentences as hypotheses. Then a part of those ``hard'' examples join the training set so as to learn a stronger model for the next round. The remaining part of ``hard'' examples act as dev/test set correspondingly. In total, three rounds were accomplished for ANLI construction. We put the data from the three rounds together as an overall dataset, which results in  train/validation/test split sizes of $162,865$/$3200$/$3200$ input pairs. 
\begin{table}[t!]
\centering
\scriptsize
\resizebox{0.47\textwidth}{!}{%
\begin{tabular}{p{7.3cm}}
\toprule
P: A man wearing a \textbf{white} shirt and an \textbf{orange} shirt jumped into the air. \\
H: A man's feet are not touching the ground. \quad\quad GT: [entailment]\\
\midrule
P: Two boxers are fighting and the one in the \textbf{purple} short is attempting to block a punch.
H: Both boxers wore \textbf{black} pants. \quad\quad GT: [contradiction]\\
P: A woman wearing a \textbf{purple} dress and \textbf{black} boots walks through a crowd drinking from a glass bottle.
H: The woman is drinking from a plastic bottle. \quad\quad GT: [contradiction]\\
P: The woman in \textbf{purple} shorts and a \textbf{brown} vest has a \textbf{black} dog to the right of her and a dog behind her. 
H: A woman is walking three dogs. \quad\quad GT: [contradiction]\\
P: A young lady wearing \textbf{purple} and \textbf{black} is running past an \textbf{orange} cone.
H: The young lady is walking calmly. \quad\quad GT: [contradiction]\\
\midrule
P: A man and woman stand in front of a large red modern statue. 
H: The man and woman are not sitting. \quad\quad GT: [entailment]\\
P: Three guys sitting on rocks looking at the scenery. H: The people are not standing. \quad\quad GT: [entailment]\\
P: Two boys are playing ball in an alley. 
H: The boys are not dancing. \quad\quad GT: [entailment]\\
P: A girl is blowing at a dandelion. H: The girl isn't swimming in a pool. \quad\quad GT: [entailment]\\
\bottomrule
\end{tabular}
}
\caption{\small The first block shows an example from the original dev set that BERT predicts incorrectly as \textit{contradiction} when trained on the SNLI original split but predicts correctly as \textit{entailment} when trained with the counterfactual augmentations. Only showing top \(k\) = 4 examples. Bold text indicates uncovered likely spurious association between mention of colors and \textit{contradiction}.}
\label{tab:nli}
\end{table}

\vskip 1em
\textbullet\enspace \textbf{HANS:}
Heuristic Analysis for NLI Systems (HANS)~\cite{mccoy2019right} is a controlled evaluation dataset aiming to probe if a model has learned the following three kinds of spurious heuristic signals: lexical overlap, subsequence, and constituent.\footnote{Available at ~\url{https://github.com/hansanon/hans}} This dataset intentionally includes examples where relying on these heuristics fail by generating from $10$ predefined templates. This dataset is challenging because state-of-the-art models like BERT~\cite{devlin2018bert} perform very poorly on it. There are in total $30,000$ examples -- $10,000$ for each heuristic. 

We use this dataset only for validating and testing our models that are trained on the ANLI dataset. The HANS dataset has only two classes, `entail' and `not-entail' while ANLI has 3 classes so we collapse the `neutral' and `contradiction' predictions into `not-entail'.  We randomly split 30K examples into 10K for validation and 20K for testing while maintaining the balance across the different heuristics in both the splits.

\subsection{System setup}
We focus on experimenting with two transformer models -- the BERT and RoBERTa. For both models we only use their base versions with 110M and 125M parameters. More details about the hyper-parameters used for each of these models are mentioned in Appendix A.

\begin{figure}[t!]
  \centering
\subfigure{
  \includegraphics[width=0.48\textwidth,height=1.5cm]{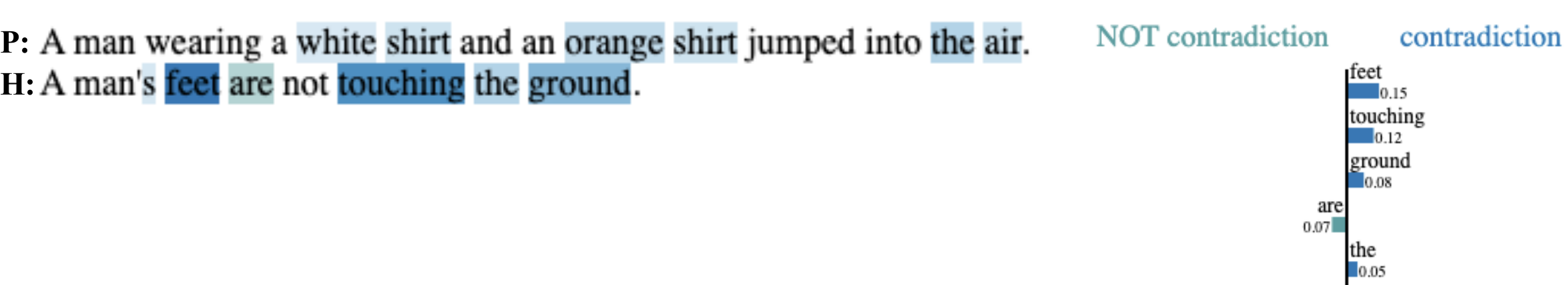}
}
\subfigure{
  \includegraphics[width=0.48\textwidth,height=1.5cm]{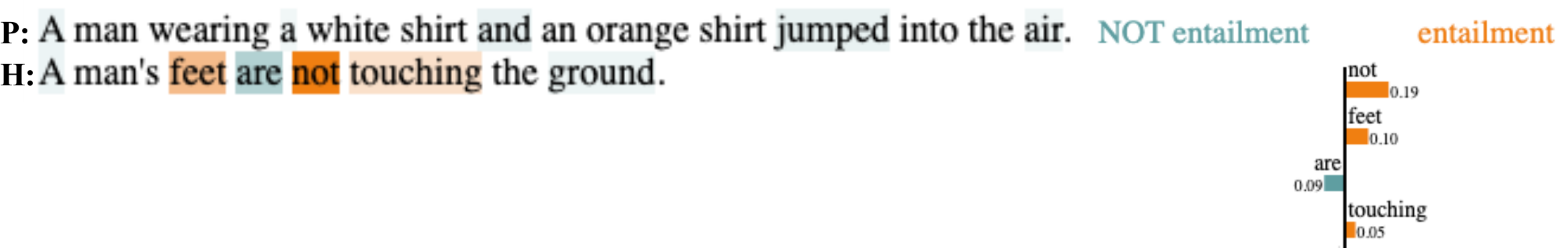}}
  \caption{\small LIME saliency map highlighting the top feature words in the BERT model's prediction. The top figure shows that the model uses the occurrence of colors as important features, verifying the spurious pattern of associating colors with the contradiction class that is uncovered by kNN in Table~\ref{tab:nli} (middle block). The bottom figure shows the contrast that \emph{after} augmentation the model does not use color features and so is able to make the right prediction (entailment). }
  \label{fig:lime}
\end{figure}

\subsection{Results}
We use \knn~as a lens into interpreting and understanding model behavior based on its learned representations. Specifically, we explore the effectiveness of \knn~on uncovering spurious associations and identifying mislabeled training examples.
In the following sections, we describe these application of \knn~using NLI as a case study.

\begin{figure}[t!]
  \centering
  \subfigure{
  \includegraphics[width=0.48\textwidth, height=1.5cm]{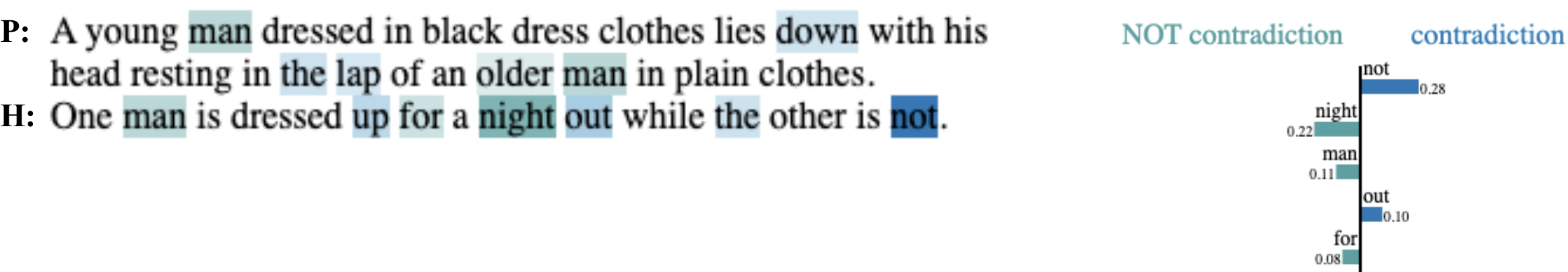}}
  \subfigure{\includegraphics[width=0.48\textwidth, height=1.6cm]{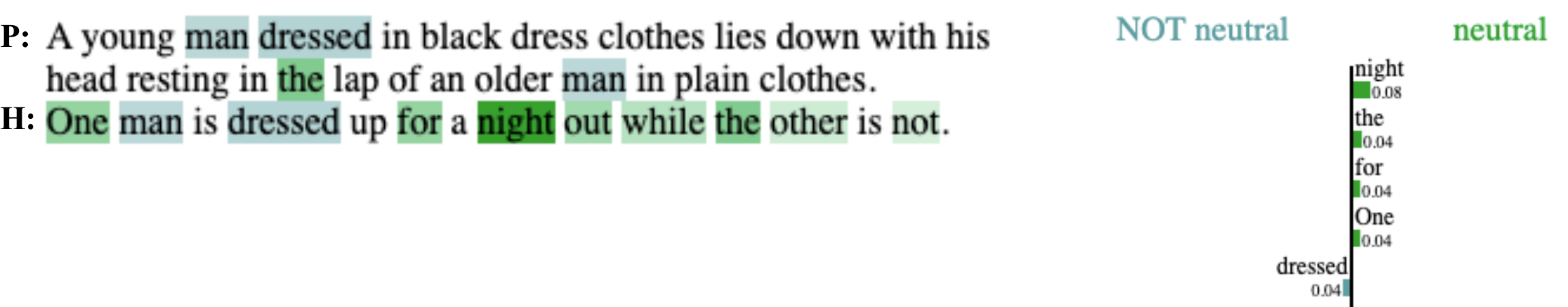}}
  \caption{\small LIME saliency map highlighting the top feature words in the BERT model's prediction. The top figure shows that the occurrence of `not' causes BERT trained on counterfactual augmentation to misclassify the example as contradiction, verifying the spurious pattern of associating negation with the contradiction class that is uncovered by kNN in Table~\ref{tab:nli-aug} (bottom block). The bottom figure shows the contrast that \emph{before} augmentation the model did not not have those spurious associations and so was able to make the right prediction (neutral).}
  \label{fig:lime2}
\end{figure}

\vskip 1em
\textbullet\enspace \textbf{Explaining the model behavior using nearest training examples.} The most similar training examples retrieved by \knn~provide context for a given input in the representation space of the model, thereby providing an understanding for why the model made a certain prediction. We run experiments to test at the dataset level if the retrieved training examples are actually the ones that the model relies on to learn its decision boundary. We do this by removing a percentage of the training examples most frequently retrieved by \knn~ (with $k=16$) on the training set, retrain the model from initialization, and re-evaluate the model. We repeat this procedure and average results over three random seeds. On the original SNLI split, we find that on average BERT's performance drops by $4.6\%$ when the top \(10\%\) of the $1666$ training examples are removed vs. $1.2\%$ when an equal amount of random examples are removed. The performance further drops by another $6\%$ when the percentage is increased to \(30\%\) vs. $4\%$ for random. 
This experiment verifies that a model's prediction for a specific testing example is highly correlated with its neighboring examples in training set. Tables~\ref{fig:lime} and ~\ref{fig:lime2} show top (based on distance) influential training examples for a dev example.
\begin{table}[t!]
\centering
\scriptsize
\resizebox{0.47\textwidth}{!}{%
\begin{tabular}{p{7.3cm}}
\toprule
P: A young man dressed in black dress clothes lies down with his head resting in the lap of an older man in plain clothes. \\
H: One man is dressed up for a night out while the other is \textbf{not}.\\ GT: [neutral]\\
\midrule
P: There is a performance with a man in a t-shirt and jeans with a woman in all black while an audience watches and laughs. \\ H: The man and woman are friends. \quad\quad GT: [neutral]\\
P:A man in a black shirt plays the guitar surrounded by drinks. \\ H: The man is skilled at playing.  \quad\quad GT: [neutral]\\\
P: An older guy is playing chess with a young boy. \\ H: Two generations play an ancient game.  \quad\quad GT: [entailment]\\
P: A young girl skiing along side an adult. \\ H: The young girl is the adult's child. \quad\quad GT: [neutral]\\
\midrule
P: Many people in white smocks look at things under identical looking microscopes. H: Multiple humans are \textbf{not} looking under microscopes. \quad\quad GT: [entailment]\\
P: A small girl dancing in a parade wearing bright red and gold clothes. H: The small girl is sitting at home, \textbf{not} celebrating. \quad\quad GT: [contradiction]\\
P: A five-piece band, four of the men in red outfits and one of them in a leather jacket and jeans, perform on the sidewalk in front of a shop. \\ H: The five-piece band  of women in red is \textbf{not} performing. \quad\quad GT: [contradiction]\\
P: Three young girls chapping and texting on a cellphone. H: Three young girls are sitting together and \textbf{not} communicating. \quad\quad GT: [contradiction]\\
\bottomrule
\end{tabular}
}
\caption{\small This table contrasts the results of \knn~when training BERT on original SNLI vs. with counterfactual augmentations. \knn~reveals artifacts in the SNLI augmented split that cause the BERT's prediction on a dev example (first block) to go from \textit{neutral} when trained on the SNLI original split (second block) to \textit{contradiction} when trained with counterfactual augmentation (third block). Only showing top \(k\) = 4 examples. Bold text indicates likely spurious correlation. Some examples have mislabeled GT but kNN is robust in identifying similar examples.}
\label{tab:nli-aug}
\end{table}

\vskip 1em
\textbullet\enspace \textbf{Uncovering spurious associations}.
Spurious associations are caused by a model confounding the statistical co-occurrence of a pattern in the input and a class label with high mutual information between the two.
 For example, state-of-the-art models are known to associate high lexical overlap between the premise and the hypothesis with the label entailment~\citep{mccoy2019right}. So models that rely on this association fail spectacularly when the subject and the object are switched. Counterfactual data augmentation alleviates this problem by reducing the co-occurrence of such artifacts and the associated class label.

\knn~provides a tool for uncovering potential spurious associations. First, we examine the nearest neighbors of misclassified examples for possible spurious patterns. Next, we use feature-importance methods like LIME to verify the pattern by comparing it to the highest-weighted word features. Table~\ref{tab:nli} shows potential spurious association between mention of colors and contradiction label uncovered by \knn~when BERT is trained on the original split. As shown, counterfactual augmentation helps in debiasing the model and BERT is then able to classify the same example correctly.

Surprisingly, through \knn~we also find that counterfactual augmentation inadvertently introduces new artifacts that degrade model performance on certain data slices. Table~\ref{tab:nli-aug} shows an example where BERT's prediction goes from neutral to contradiction when trained on augmented data. The nearest neighbors reveal that BERT learns to correlate the occurrence of negation in the hypothesis with contradiction. LIME verifies that the most highly weighted feature is the occurrence of `not' as shown in Figure~\ref{fig:lime2}. Quantitatively, we verify that the pattern `\textit{not}' occurs approximately $37\%$ and $60\%$ of times in the original and augmented training splits of SNLI respectively. The accuracy of BERT on identifying entailment examples that contain negation drops by $10\%$ when trained with augmented data versus without. Figure~\ref{fig:lime2} shows the saliency map of words that were highly-weighted in BERT's prediction using LIME. As expected, the model trained on augmented data learns to spuriously associate the occurrence of `\textit{not}' with the contradiction class.

\vskip 1em
\textbullet\enspace \textbf{Identifying mislabeled examples}.
Datasets used for training deep learning models are huge and may contain noisy labels depending on how the data was collected. Even crowd-sourced datasets can sometimes have mislabeled examples~\cite{frenay2013classification}. \Knn~can be leveraged for identifying potentially mislabeled examples. We observed that \knn~would sometimes retrieve mislabeled examples, and that these specific instances tended to occur when \knn's prediction was different from the models prediction. 

We experimented with this phenomenon by intentionally mislabeling 10\% of examples on the original training split of the SNLI dataset with 1666 examples, and using \knn~ based method to recover them by comparing with the model's prediction as follows (based on the notations used in Section~\ref{sec:model}):
\[
\argmax p_{\theta}(y|x') \neq \argmax p_{knn}(y|x')
\]
We collected a set of candidate mislabeled training examples by comparing BERT's prediction on the dev set to the label of the immediate nearest neighbor (\(k\)=1) for that example. We found that our approach was extremely effective at identifying the mislabeled examples with both high precision and recall. We obtained precision, recall and F1 of $0.84$, $0.73$, and $0.78$ respectively averaged over three random seeds.

\begin{figure}[ht!]
  \centering
  \includegraphics[width=0.45\textwidth]{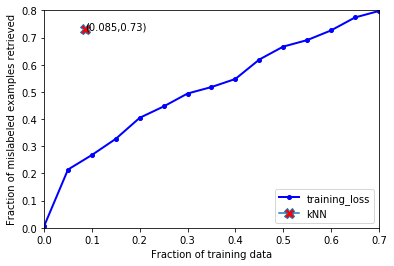}
  \caption{\small Fraction of correctly retrieved  mislabeled examples by our \knn~approach compared to a baseline that considers examples with highest training loss as mislabeled. Our approach has high recall while retrieving a significantly smaller set of candidate mislabeled examples.}
  \label{fig:mislabel}
\end{figure}

We compared our results to a baseline that classifies training examples with the highest training loss as potentially mislabeled. Figure~\ref{fig:mislabel} shows a plot of the baseline with respect to the fraction of training data and the corresponding recall. Because our approach retrieves a set of candidate examples and there is no ranking, we directly plot the performance based on the size of the set. Our results indicate that \knn~is extremely effective at identifying the mislabeled examples compared to the baseline that requires about 65\% of the training data ranked by loss to get to the same performance. We applied our approach to the counterfactually augmented split of SNLI and found that our method effectively uncovers several mislabeled examples as shown in Table~\ref{tab:misalbel}.

\begin{table}[t!]
\centering
\scriptsize
\resizebox{0.47\textwidth}{!}{%
\begin{tabular}{p{7.3cm}}
\toprule
P: A person uses their laptop. 	H: A person uses his laptop in his car.	\quad\quad GT: [entailment]\\
P: A man in a white shirt at a stand surrounded by beverages and lots of lemons. H: The man is selling food.	\quad\quad GT: [contradiction]\\
P: Extreme BMX rider with no gloves and completing a jump.	H: A bike rider has biking gloves. \quad\quad 	GT: [contradiction]
\\
P: A family is observing a flying show.	H: A family looking up into the sky at a flight show outside.	\quad\quad GT: [entailment]
\\
P: A man is riding a motorcycle with a small child sitting in front of him. H: A man rides his motorcyle with his won. \quad\quad GT: [entailment]\\
P: A man is riding a blue truck with a small child sitting in front of him. H: A man rides his motorcyle with his won. \quad\quad GT: [contradiction]\\
P: A girl jumping into a swimming pool. H: A girl is taking a swim outside.	\quad\quad GT: [entailment]\\
P: A young woman wearing a black tank top is listening to music on her MP3 player while standing in a crowd. H: A young woman decided to leave home.\quad\quad GT: [entailment]\\
\bottomrule
\end{tabular}
}
\caption{\small Random sample of retrieved candidate mislabeled examples from the SNLI augmented split. Not all are mislabeled and some also have mispelled words.}
\label{tab:misalbel}
\end{table}

\begin{figure*}[ht!]
  \centering
  \includegraphics[width=\textwidth]{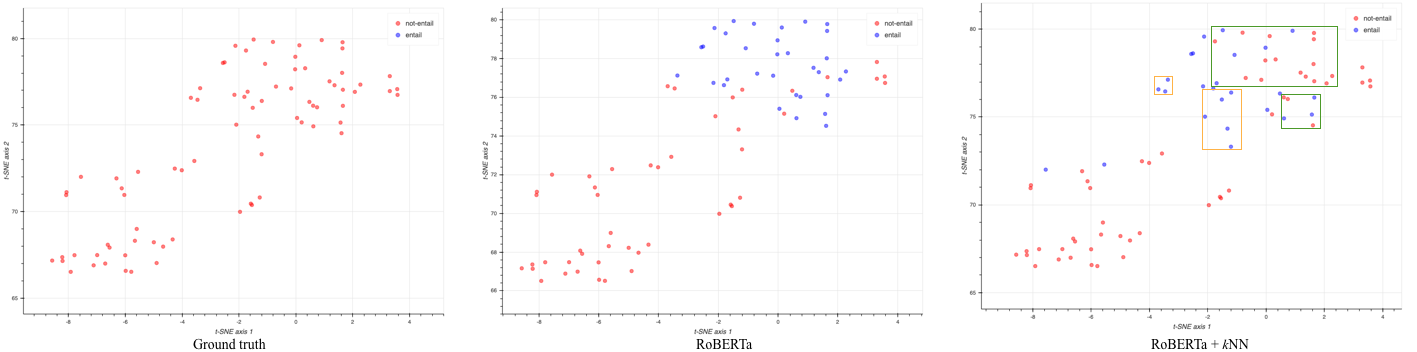}
  \caption{\small Visualization of RoBERTa's representation using t-SNE on a small subset of the HANS constituent heuristic portion of the val set. Because \knn~is highly non-linear, it is able to recover from some of the misclassifications of RoBERTa (highlighted using green box) for the difficult non-entail class. It makes the performance slightly worse on much smaller clusters (highlighted using orange box).}
  \label{fig:tsne}
\end{figure*}
Apart from explaining model behavior and identifying mislabeled examples, we explore mechanisms for leveraging \knn~to further improve fine-tuned model predictions described in the next section.

\vskip 1em
\textbullet\enspace \textbf{\Knn~for improving model performance}.
\Knn~has the ability to learn a highly non-linear boundary and so we leverage it to improve performance of fine-tuned models on examples which we know the model is not good at classifying. We deploy \knn~as a backoff for low confidence predictions by learning a threshold on the validation set, below which the model's predictions are very unreliable. Another criteria could be defining slices of data that satisfy a property on which the model is known to perform poorly. Examples include inputs that contain gendered words or fallible patterns for a model.
\begin{table}[t!]
\centering
\resizebox{0.49\textwidth}{!}{%
\begin{tabular}{lS[table-format=1.3]S[table-format=1.3]S[table-format=1.3]S[table-format=1.3]}
\toprule
\multirow{2}{*}{} & \multicolumn{2}{c}{SNLI} &\centering{ANLI}& HANS   \\ 
\cmidrule{2-3}
&{Original}&{Augmented}&&\\
&\(k\)=16 &\(k\)=16&\(k\)=64&\(k\)=64\\
\midrule
BERT       &  75.7&70.5&47.2&52.4\\
\(k\)NN- BERT &75.8&71.0&48.5&52.4\\
\midrule
RoBERTa       &82.8&76.8&44.4&59.3\\
\(k\)NN-RoBERTa   &82.8  &77.3&47.9&59.6\\
\bottomrule
\end{tabular}
}
\caption{\small Results on NLI test sets for BERT and RoBERTa compared to \knn~used along with the underlying model. Numbers indicate accuracy. SNLI and ANLI are 3-way classification tasks while HANS is a 2-way classification task. }
\label{tab:results_salient}
\end{table}

\paragraph{Results on NLI.}
For all our experiments we used \knn~as a backoff for examples where the underlying fine-tuned model has low confidence in its prediction. These thresholds are identified on the validation splits for each of the evaluation datasets. We also experimented with different values of \(k\) ($<1\%$ of the size of training data) and the temperature T for different training data sizes. 

\begin{table}[t!]
\centering
\begin{tabular}{@{}lrr@{}}
\toprule
Dataset&BERT&RoBERTa\\
\midrule
SNLI original&0.74&0.55\\
SNLI augmented&0.53&0.40\\
ANLI&0.95&0.96\\
HANS&0.84&0.62\\
\bottomrule
\end{tabular}
\caption{Threshold $\tau$ used for switching to \knn~identified on the validation sets.}
\label{tab:knn_param}
\end{table}

Table~\ref{tab:results_salient} shows the performance of BERT and RoBERTa with and without \knn~on the overall test set for three NLI datasets. Our method of combining \knn~with the underlying model obtains beats the standard BERT/RoBERTa with big margins on both augmented SNLI and ANLI. We also observe slight improvements on some of the other datasets.

To get a better insight into how \knn~improves the fine-tuned models, we visualized RoBERTa's learned representations over a sample of the HANS validation set. The sample is chosen from the particularly difficult \emph{constituent} heuristic of HANS that assumes that a premise entails all complete sub-trees in its parse tree~\citep{mccoy2019right}.
Figure~\ref{fig:tsne} illustrates how the predictions change when using just fine-tuned RoBERTa vs. in combination with \knn. Our approach does particularly well on the more difficult \emph{not-entail} class as identified by~\citet{mccoy2019right}. Fine-grained quantitative results shows in Table~\ref{tab:hans}. We used t-SNE~\citep{maaten2008visualizing} to project the representations into two dimensions for visualization. 

\begin{table}[t!]
\centering
\resizebox{0.49\textwidth}{!}{%
\begin{tabular}{@{}lrrr@{}}
\toprule
&Lexical overlap&Subsequence&Constituent\\
\midrule
BERT       &23.3 &30.6&18.2\\
\(k\)NN- BERT &25.4&30.6&24.6\\
\midrule
RoBERTa       &78.7&79.6&52.7\\
\(k\)NN-RoBERTa   &79.4&80.8&54.7\\
\bottomrule
\end{tabular}
}
\caption{\small Fine-grained performance on HANS test set for the more difficult \emph{not-entail} class. \knn~consistently outperforms the underlying model by itself.}
\label{tab:hans}
\end{table}

\paragraph{Analysis.}
Our experiments indicate that \knn~can improve the performance of state-of-the-art models especially on input types on which the model is known to perform poorly. In this paper, we only considered the model's low confidence as an indicator for switching to \knn. The backoff criteria could be anything that is based on the input examples. Slicing the datasets based on the occurrence of certain patterns in the input text like mention of colors or criterion based on syntactic information such as part-of-speech tags or lexical overlap can give a deeper understanding of model behavior. Fine-grained evaluations on such slices on a validation set would highlight data slices where the model performs poorly. Example types that satisfy these criterion can then be classified by switching to \knn~for the final prediction.

Based on our application of \knn~for uncovering spurious associations, it is evident that high performance on data augmentation such as counterfactuals that are not targeted for pre-defined groups or patterns in examples does not indicate robustness on all slices of data (e.g. examples with the `not' pattern). It is difficult to make any claims about robustness without performing fine-grained analysis using data slices. On the other hand, targeted augmentation techniques such as HANS work better for evaluating robustness on the pre-defined syntactic heuristic categories.
\section{Conclusion and future directions}
We leveraged \knn~over hidden representations to study the behavior of NLI models, and find that this approach is useful both for interpretability and improving model performance. We find that the \knn~for any test example can give useful information about how a model makes its classification decisions. This is especially valuable for studying where models go wrong, and we demonstrate how retrieving the nearest neighbors can reveal artifacts and spurious correlations that cause models to misclassify examples. 

From our analysis, we also observed that \knn~of misclassified test examples are often indicative of mislabeled examples, which gives this approach application to fixing mislabeled examples in training sets. By finding the most common nearest neighbors across the whole test set, we are also able to identify a subset of highly influential training examples and obtain corpus-level interpretability for the model's performance. 

Lastly, we examined the utility of backing off to \knn-based classification when the model confidence is low. Analysis of the decision boundaries learned by \knn~over hidden representations suggest that kNN learns a more fine-grained decision boundary that could help make it more robust to small changes in text that cause the ground truth label, but not the model prediction, to flip. We find that this approach increases the robustness of models to adversarial NLI datasets.

Future work could extend the effectiveness of \knn~to other applications such as identifying domain mismatch between training and test or finding the least influential examples in a training set. Other directions include extending \knn~to other classification tasks as well as generation.

\appendix

\bibliography{anthology,aaai21}

\end{document}